\documentclass[letterpaper]{article} 
\usepackage{aaai24}  
\usepackage{times}  
\usepackage{helvet}  
\usepackage{courier}  
\usepackage[hyphens]{url}  
\usepackage{graphicx} 
\usepackage{natbib}  
\usepackage{caption} 
\usepackage{algorithm}
\usepackage{algorithmic}

\usepackage[table]{xcolor}
\usepackage{amsmath}
\usepackage{multirow}
\usepackage{pifont}

\usepackage{newfloat}
\usepackage{listings}
\DeclareCaptionStyle{ruled}{labelfont=normalfont,labelsep=colon,strut=off} 
\floatstyle{ruled}
\newfloat{listing}{tb}{lst}{}
\floatname{listing}{Listing}
\title{Hybrid-SORT: Weak Cues Matter for Online Multi-Object Tracking}
\author{
    Mingzhan Yang\textsuperscript{\rm 1,\rm 2}\equalcontrib,
    Guangxin Han\textsuperscript{\rm 1}\equalcontrib,
    Bin Yan\textsuperscript{\rm 1},
    Wenhua Zhang\textsuperscript{\rm 1},\\
    Jinqing Qi\textsuperscript{\rm 1},
    Huchuan Lu\textsuperscript{\rm 1},
    Dong Wang\textsuperscript{\rm 1}\thanks{Corresponding authors.}
}
\affiliations{
    \textsuperscript{\rm 1} Dalian University of Technology\\
    \textsuperscript{\rm 2} Shenzhen Tvt Digital Technology Co., Ltd\\
    \{18742525689, hanguangxindlut, yan\_bin, 1062894314zwh\}@mail.dlut.edu.cn, \{jinqing, lhchuan, wdice\}@dlut.edu.cn
}

\usepackage{bibentry}

\begin{document}

\maketitle

\begin{abstract}
Multi-Object Tracking (MOT) aims to detect and associate all desired objects across frames. Most methods accomplish the task by explicitly or implicitly leveraging strong cues (i.e., spatial and appearance information), which exhibit powerful instance-level discrimination. However, when object occlusion and clustering occur, spatial and appearance information will become ambiguous simultaneously due to the high overlap among objects. In this paper, we demonstrate this long-standing challenge in MOT can be efficiently and effectively resolved by incorporating weak cues to compensate for strong cues. Along with velocity direction, we introduce the confidence and height state as potential weak cues. With superior performance, our method still maintains Simple, Online and Real-Time (SORT) characteristics. Also, our method shows strong generalization for diverse trackers and scenarios in a plug-and-play and training-free manner. Significant and consistent improvements are observed when applying our method to 5 different representative trackers. Further, with both strong and weak cues, our method Hybrid-SORT achieves superior performance on diverse benchmarks, including MOT17, MOT20, and especially DanceTrack where interaction and severe occlusion frequently happen with complex motions. The code and models are available at https://github.com/ymzis69/HybridSORT.
\end{abstract}

\section{Introduction}

\begin{figure}[!t]
\centering
\includegraphics[width=0.95\columnwidth]{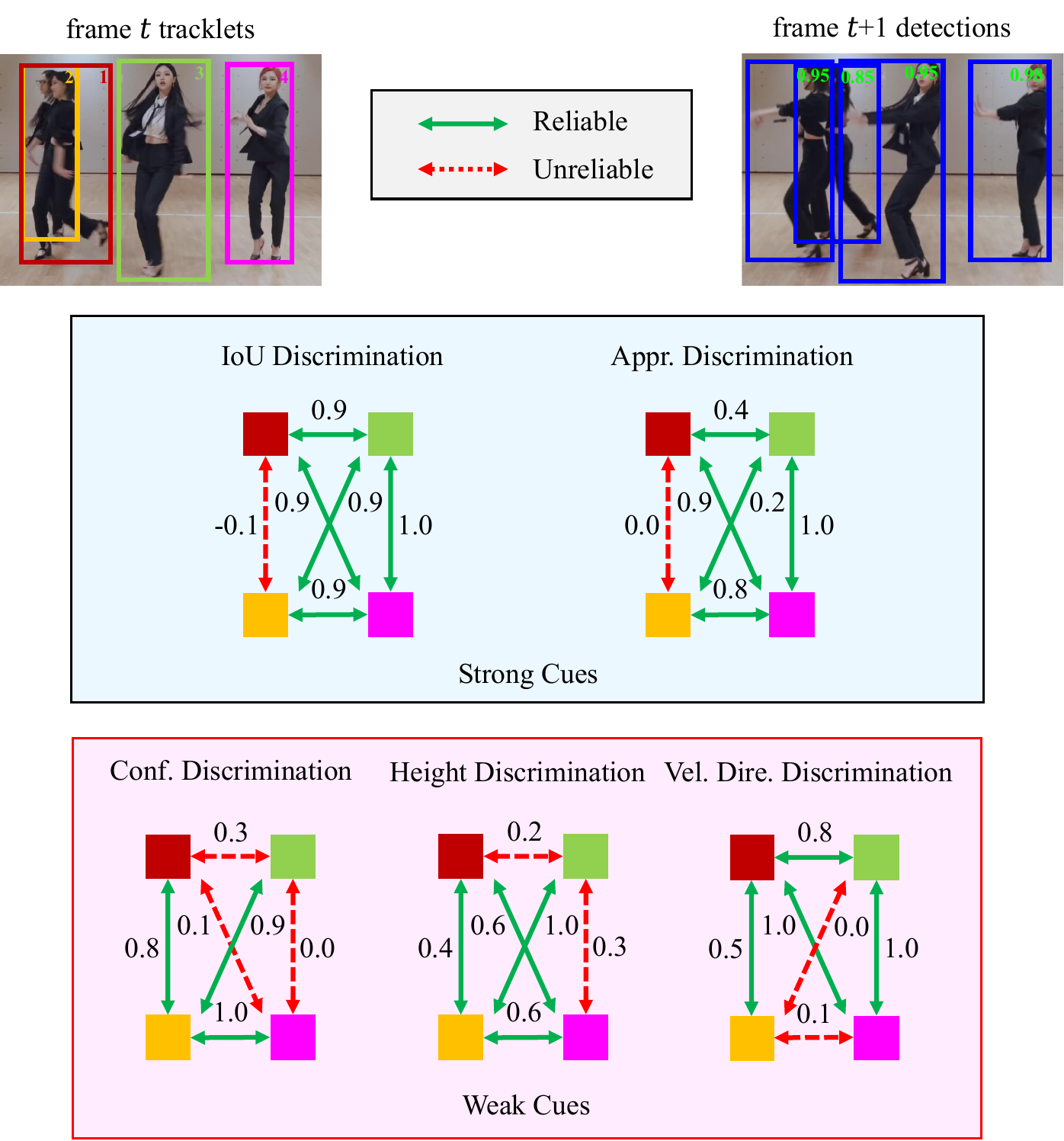} 
\caption{The discrimination capacity of strong and weak cues. Green solid arrows represents reliable discrimination between pairwise objects, while red dashed arrows indicate unreliable discrimination. The higher the value of the arrow, the more reliable the discrimination is.} 
\label{Motivation}
\end{figure}

Recently, tracking-by-detection \cite{bewley2016simple, wojke2017simple, zhang2021fairmot, zhang2022bytetrack, du2023strongsort, ren2023focus, cao2023observation} has become the most popular paradigm in Multi-Object-Tracking (MOT), which divides the problem into two sub-tasks. The first task is to detect objects in each frame. The second task is to associate them in different frames. The association task is primarily solved by explicitly or implicitly utilizing strong cues, including spatial and appearance information. This design is reasonable because these strong cues provide powerful instance-level discrimination. However, the commonly used strong cues suffer from degradation under challenging situations such as occlusion and clustering (ID 1 and 2 in Figure \ref{Motivation}). Specifically, when two objects are highly overlapped in the current frame, the Intersection over Union (IoU) between detections and estimated tracklet locations becomes ambiguous, and the appearance features of both objects are dominated by the foreground ones (red dash arrow in the \textit{Strong Cues} part of Figure \ref{Motivation}).

In the \textit{Weak Cues} part of Figure \ref{Motivation}, we demonstrate that weak cues, such as confidence state, height state, and velocity direction, can effectively alleviate the ambiguous associations where strong cues become unreliable. However, to the best of our knowledge, weak cues have been ignored in most methods except for very few (e.g., OC-SORT \cite{cao2023observation}, MT-IOT \cite{yan2022multiple}), as they only possess reliable discrimination among certain objects. As shown in Figure \ref{Motivation}, the confidence state is only reliable for distinguishing ID 2 from other IDs. 

In this paper, we select the confidence state and height state as potential types of weak cues, in addition to the velocity direction used in OC-SORT \cite{cao2023observation}. The confidence state can explicitly indicate the occluding/occluded (i.e., foreground/background) relations among clustered objects, providing a critical clue that strong cues (i.e., spatial and appearance information) lack. Height state is a stable property of objects which is usually robust to diverse object poses and contains some degree of depth information (i.e., reflects the distance from the camera to the objects).

To maintain the Simple, Online and Real-Time (SORT) characteristics, we propose simple yet effective strategies to exploit the aforementioned weak cues, namely Tracklet Confidence Modeling (TCM) and Height Modulated IoU (HMIoU). For TCM, we use Kalman Filter and Linear Prediction to estimate the confidence state of tracklets, which is then used as a metric to associate with detections. For HMIoU, the height state is also modeled by Kalman Filter. The height cost matrix for the association is first defined as the IoU along the height axis for the estimated tracklet box and detection box, then fused with the standard IoU matrix based on the area metric.

To evaluate the generalization ability of our design, we apply the proposed designs to 5 different representative trackers, including SORT \cite{bewley2016simple}, DeepSORT \cite{wojke2017simple}, MOTDT \cite{chen2018real}, ByteTrack \cite{zhang2022bytetrack}, and OC-SORT \cite{cao2023observation}. Both of our designs for confidence state and height state consistently achieve significant improvements, demonstrating the importance of weak cues for MOT.

Further, to advance the state-of-the-art performance of Simple, Online and Real-Time (SORT) MOT methods, we modify the current state-of-the-art SORT-like algorithm OC-SORT \cite{cao2023observation} as our strong baseline. Firstly, we modify the velocity direction modeling in OC-SORT, namely Observation-Centric Momentum (OCM), by extending the box center to four box corners and the fixed temporal interval to multiple intervals. Secondly, we include an additional association stage for low-confidence detection following ByteTrack \cite{zhang2022bytetrack}. Along with the proposed TCM and HMIoU, our method Hybrid-SORT achieves superior performance on all DanceTrack, MOT17, and MOT20 benchmarks by leveraging both strong and weak cues, while still maintaining Simple, Online and Real-Time (SORT). We hope that the generalization ability, plug-and-play and training-free characteristics of Hybrid-SORT make it attractive for diverse scenarios and edge devices.

\begin{itemize}
\item We demonstrate the long-standing challenges of occlusion and clustering in MOT can be substantially alleviated by incorporating weak cues (i.e., confidence state, height state and velocity direction) as compensation for commonly used strong cues.
\item We introduce simple Tracklet Confidence Modeling (TCM) and Height Modulated IoU (HMIoU) to model and leverage the confidence state and height state. With delicate modeling, the weak cues effectively and efficiently relieve the ambiguous matches generated by strong cues with negligible additional computation.
\item The plug-and-play and training-free design generalizes well over diverse scenarios and trackers. We implement our design on 5 representative trackers, achieving consistent and significant improvements. Finally, Our method Hybrid-SORT achieves superior performance on DanceTrack, MOT17, and MOT20 benchmarks.
\end{itemize}

\section{Related Work}
\subsection{Heuristic Matcher}
\subsubsection{Spatial-based Heuristic Matcher}
Spatial information is the most widely used strong cue in high-FPS benchmarks. When time intervals between frames are short, the movement of an object is also small and can be treated as linear. This makes spatial information an accurate metric in the short-term association. The pioneer work SORT \cite{bewley2016simple} uses Kalman Filter \cite{kalman1960contributions} to predict the spatial locations of tracklets and perform associates based on the IoU metric. Subsequent works, such as CenterTrack \cite{zhou2020tracking}, ByteTrack \cite{zhang2022bytetrack}, MotionTrack \cite{qin2023motiontrack}, and OC-SORT \cite{cao2023observation}, are all heuristic matching that only utilize spatial information for association. However, even the most advanced method, OC-SORT \cite{cao2023observation}, still suffers from heavy occlusion and clustering.

\subsubsection{Appearance-based Heuristic Matcher}
Unlike spatial information, appearance information possesses relatively stable consistency throughout the whole video, thus benefiting long-term association. Following SORT, DeepSORT \cite{wojke2017simple} and GHOST \cite{seidenschwarz2023simple} utilize an independent ReID model to extract appearance features for the association. Then the following work JDE \cite{wang2020towards}, FairMOT \cite{zhang2021fairmot}, CSTrack \cite{liang2022rethinking}, QDTrack \cite{pang2021quasi}, FineTrack \cite{ren2023focus} and UTM \cite{you2023utm} integrated the detection and ReID models for joint training and designed improved network architectures to enhance performance. However, we observe that among clustered objects, both spatial and appearance cues suffer from severe discrimination degradation, even if delicate network architectures and association strategies are designed. 

\subsection{Learnable Matcher}
\subsubsection{Graph-based Learnable Matcher}
Graph-based learnable matchers formulate the association task as an edge classification task, where the edge label is 1 for tracklet nodes and detection nodes with the same ID and vice versa. MOTSolv \cite{braso2020learning} and GMTracker \cite{he2021learnable} are based on Graph Neural Network (GNN) and make the data association step differentiable.  Most recently, SUSHI \cite{cetintas2023unifying} leverages graph models to hierarchically connect short tracklets into longer tracklets in an offline fashion. However, the major limitation of graph-based matchers is that the training and inference pipeline is often complicated or even offline, which restricts their practical use in online tracking scenarios that impose strict real-time demands, such as autonomous driving.

\begin{figure*}[ht]
\centering
\includegraphics[width=0.90\textwidth]{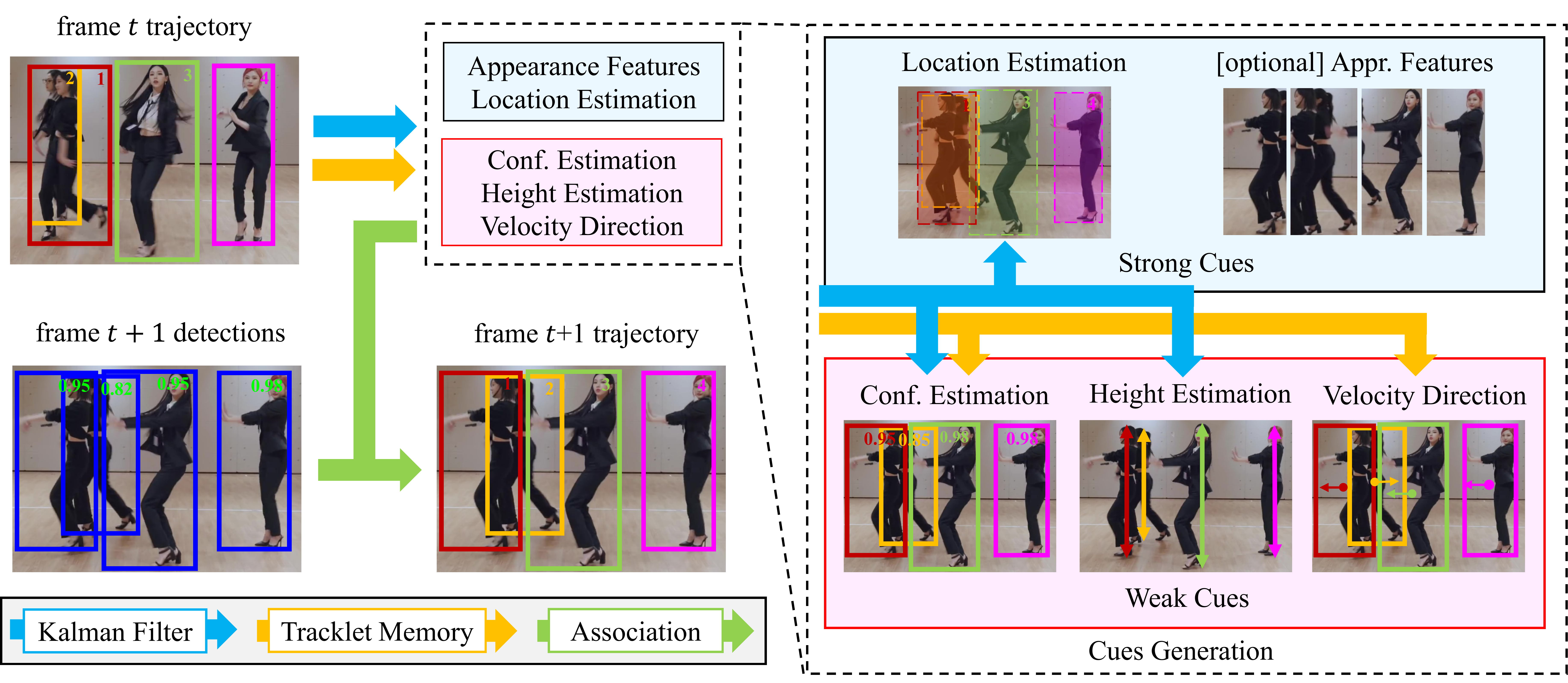} 
\caption{Pipeline of Hybrid-SORT and Hybrid-SORT-ReID. For strong cues, we utilize IoU as the metric for spatial information, and utilize cosine distance for appearance features. For weak cues, we incorporate the confidence state, height state, and velocity direction. Velocity direction is illustrated by centers instead of corners for better clarity.}
\label{pipeline}
\end{figure*}

\subsubsection{Transformer-based Learnable Matcher}
Since the Transformer became popular in vision tasks, many works are proposed to utilize its powerful attention mechanism to model the association task. TrackFormer \cite{meinhardt2022trackformer} and MOTR \cite{zeng2022motr} utilize both track queries and standard detection queries to jointly perform trajectory propagation and initialization. Most recently, MOTRv2 \cite{zhang2023motrv2} introduces a separate detector to MOTR, trying to resolve the conflict between detection and association. However, the Transformer-based matchers involve a significant number of self-attention and cross-attention operations, preventing the algorithm from achieving real-time capability.

\section{Method}
Hybrid-SORT and Hybrid-SORT-ReID follow the SORT paradigm, which utilizes Kalman Filter for motion estimation of tracklets with or without ReID module for appearance modeling. The association task is solved by Hungarian algorithm as bipartite graph matching. The cost matrices for Hungarian algorithm are computed by measuring the pairwise representation similarity between tracklets and detections. The association pipeline is shown in Figure \ref{pipeline}.

\subsection{Weak Cues Modeling}

\subsubsection{Tracklet Confidence Modeling}

\begin{figure}[ht]
\centering
\includegraphics[width=0.95\columnwidth]{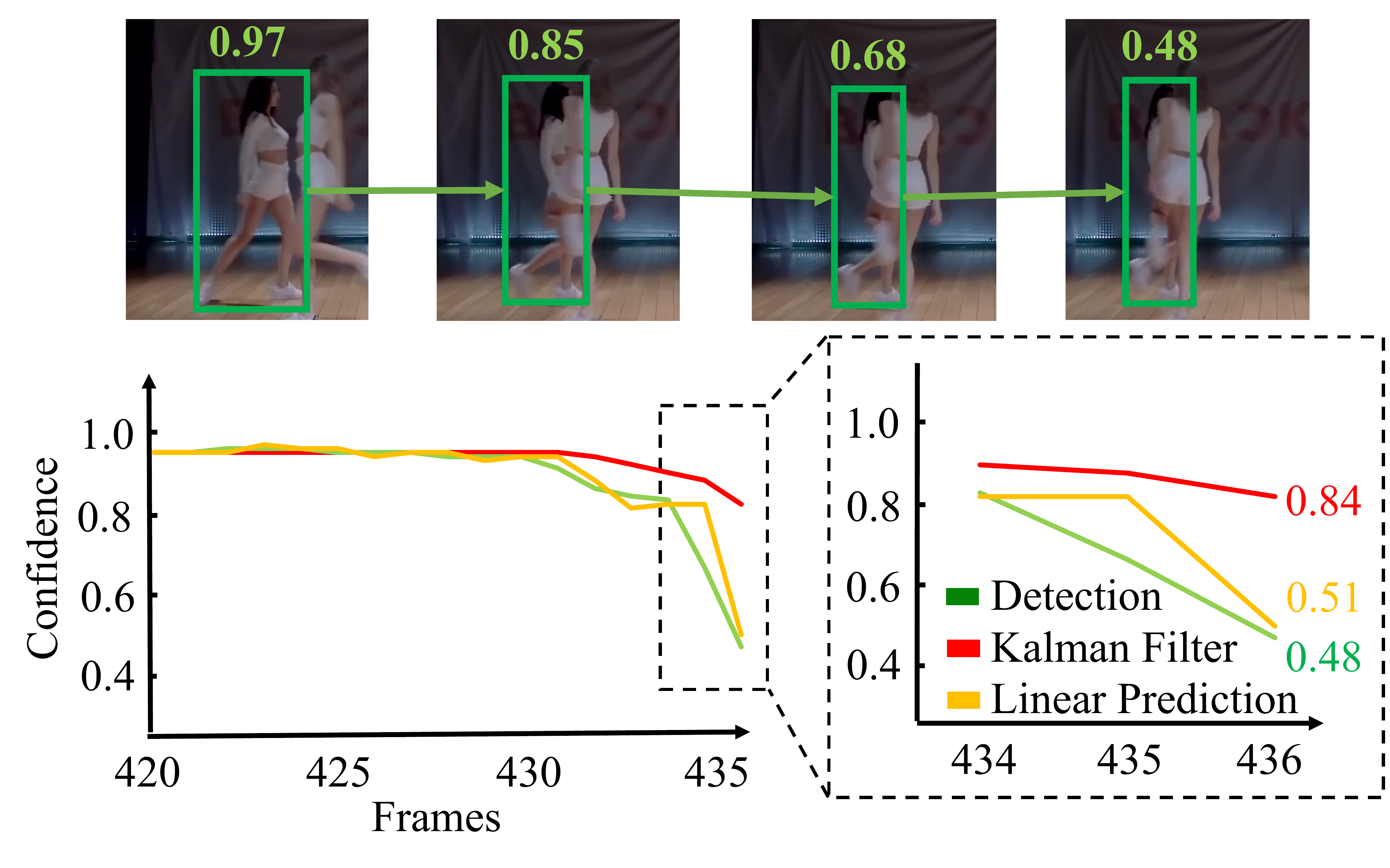} 
\caption{The confidence curve of an object. Kalman Filter estimation lags behind the actual confidence during occlusion while Linear Prediction performs effectively.}
\label{TCM}
\end{figure}

The reason why the confidence state helps association is straightforward. Specifically, when both commonly used strong cues (i.e., spatial and appearance information) fails as multiple objects are highly overlapped, the confidence of objects provides explicit foreground/background (i.e., occluding/occluded) relationships, which is exactly what strong cues lack.

Based on this insight, we introduce two modeling approaches for tracklet confidence to association with high-confidence and low-confidence detections. When objects are unobstructed or only slightly occluded, Kalman Filter is an ideal model for modeling and estimating the continuous state. Therefore, we extend the widely used standard Kalman Filter in SORT \cite{bewley2016simple} with two additional states: the tracklet confidence $c$ and its velocity component $\dot{c}$. For better clarity, we first revisit the standard Kalman Filter states in SORT, depicted in Eq. \ref{eq:sort_kf}. Here, $u$ and $v$ denote the object's center, while $s$ and $r$ represent the object box's scale (area) and aspect ratio, respectively. The velocity components are denoted by $\dot{u}$, $\dot{v}$, and $\dot{s}$.

\begin{equation}
\label{eq:sort_kf}
x=[u,v,s,r,\dot{u},\dot{v},\dot{s}]
\end{equation}

With the two newly introduced states $c$ and $\dot{c}$, the complete states of Kalman Filter in TCM are shown in Eq. \ref{eq:tcm_kf}.

\begin{equation}
\label{eq:tcm_kf}
x=[u,v,s,c,r,\dot{u},\dot{v},\dot{s}, \dot{c}]
\end{equation}

For low-confidence detections in the second association step, we utilize Linear Prediction to estimate the tracklet confidence. The confidence of objects will rapidly increase or decrease during the occlusion starts or ends. Unfortunately, Kalman Filter exhibits significant lag when attempting to estimate sudden changes in the confidence state, as shown in Figure \ref{TCM}. However, we observe clear directionality in the trend of confidence changes during this short period. Therefore, we use a simple Linear Prediction based on trajectory history to address this issue. The formula for linear modeling is given by Eq. \ref{eq:tcm_lp}, where $c_{trk}$ represents the confidence of tracklets saved in Tracklet Memory.

\begin{equation}
\label{eq:tcm_lp}
\hat{c}_{trk}=\begin{cases}
c^{t-1}_{trk}, & c^{t-2}_{trk}=None \\
c^{t-1}_{trk} - (c^{t-2}_{trk} - c^{t-1}_{trk}), & \text{else}
\end{cases}
\end{equation}

When utilizing either Kalman Filter or Linear Prediction, the confidence cost is calculated as the absolute difference between the estimated tracklet confidence $\hat{c}_{trk}$ and detection confidence $c_{det}$ following Eq. \ref{eq:confidence_cost}.

\begin{equation}
\label{eq:confidence_cost}
C_{Conf}= \lvert \hat{c}_{trk} - c_{det}\rvert
\end{equation}

\subsubsection{Height Modulated IoU}

Identifying the temporally stable properties of objects is one of the most critical aspects of multiple object tracking (MOT). The height state can provide informative clues that help to compensate for the discrimination of strong cues. Specifically, height state enhances association in two aspects. Firstly, the height of objects reflects depth information to some extent. For datasets such as DanceTrack, the heights of detection boxes mainly depend on the distance between objects and the camera. This makes the height state an effective cue for distinguishing highly overlapped objects. Secondly, the height state is relatively robust to diverse poses, making it an accurately estimated state and a high-quality representation of objects.

Specifically, we define the two boxes as $b^1 = (x_1^1, y_1^1, x_2^1, y_2^1)$ and $b^2 = (x_1^2, y_1^2, x_2^2, y_2^2)$ in which $x_1$ and $y_1$ represents the top-left corner while $x_2$ an $y_2$ represents the bottom-right corner. Also, we define the areas of two boxes as $A$ and $B$. The computation of conventional IoU is shown in Eq. \ref{eq:IoU}, which is based on the area metric. Further, the Height IoU (HIoU) can be generated by computing the IoU based on the height metric, as in Eq. \ref{eq:HIoU}.

\begin{equation}
\label{eq:IoU}
IoU = \frac{ \lvert A \cap B \rvert }{ \lvert A \cup B \rvert }
\end{equation}

\begin{equation}
\label{eq:HIoU}
HIoU = \frac{\text{min}(y_2^1, y_2^2) - \text{max}(y_1^1, y_1^2)}{\text{max}(y_2^1, y_2^2) - \text{min}(y_1^1, y_1^2)}
\end{equation}

To better utilize the height state, we introduce Height Modulated IoU (HMIoU) by combining Height IoU (HIoU) with the conventional IoU, as shown in Eq. \ref{eq:HMIoU}. The $\cdot$ means element-wise multiplication. Considering the HIoU represents the height state which is a weak cue, and IoU represents the spatial information which is a strong cue, we use HIoU to modulate the IoU by element-wise multiplication, achieving enhanced discrimination for clustered objects. 

\begin{equation}
\label{eq:HMIoU}
HMIoU = HIoU \cdot IoU
\end{equation}

\subsection{Hybrid-SORT}

\subsubsection{Robust Observation-Centric Momentum}

In OC-SORT, the Observation-Centric Momentum (OCM) considers the velocity direction of object centers in the association. The cost metric used in OCM is the absolute difference between the tracklet velocity direction $\theta_t$ and the tracklet-to-detection velocity direction $\theta_d$ in radians format, which is expressed as $\Delta \theta = \lvert \theta_t - \theta_d \rvert$. The tracklet velocity direction is obtained from two box centers in the tracklet at a temporal interval $\Delta t$, and the tracklet-to-detection velocity direction is obtained from the centers of a tracklet historical box and a new detection box. Given two points $(u_1, v_1)$ and $(u_2, v_2)$, the velocity direction is computed as Eq. \ref{eq:velocity_direction}. However, the modeling of the original OCM is vulnerable to noise caused by fixed temporal intervals and sparse points (i.e., only object centers).

\begin{equation}
\label{eq:velocity_direction}
\theta = \text{arctan} \big( \frac{v_1-v_2}{u_1-u_2} \big)
\end{equation}

\begin{figure}[t]
\centering
\includegraphics[width=0.9\columnwidth]{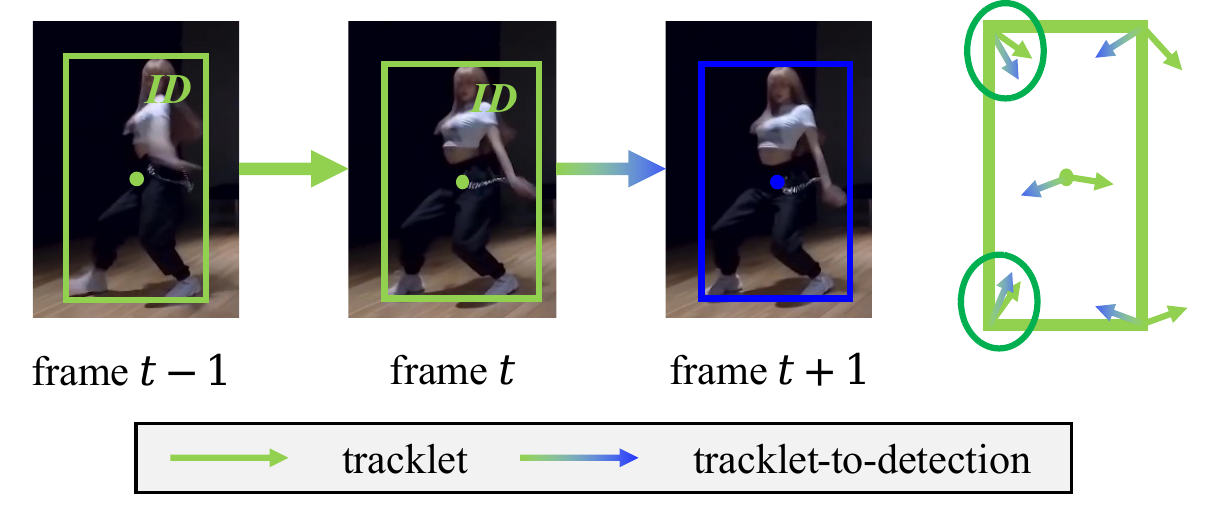} 
\caption{Velocity direction of the center and corners. While the velocity direction of some corners maintains high similarity, the direction of the center is completely opposite.}
\label{Better_OCM}
\end{figure}

We improve the OCM by introducing more robust modeling of the velocity direction, namely Robust Observation-Centric Momentum (ROCM). The modifications include two aspects. Firstly, we extend the fixed time interval of 3 frames to the stack of multiple intervals ranging from 1 to 3. Secondly, we use the four corners of the object instead of its center point to calculate the velocity direction. With multiple temporal intervals and points, the calculation formula for the ROCM is as Eq. \ref{eq:ROCM}. Figure \ref{Better_OCM} illustrates that for objects with complex motions, the velocity direction of corners maintains high similarity, while the direction of the center is nearly opposite.

\begin{equation}
\label{eq:ROCM}
C_{Vel}= \sum_{\Delta t=1}^3\frac{(C_{\Delta t}^{lt}+C_{\Delta t}^{rt}+C_{\Delta t}^{lb}+C_{\Delta t}^{rb})}{4}
\end{equation}

\subsubsection{Appearance Modeling}
We incorporate appearance information using an independent ReID model, as illustrated in Figure \ref{Motivation}. Following BoT-SORT, our pipeline first detects objects and then feeds the resulting cropped patches into the ReID model. We model tracklet appearance information using Exponential Moving Average (EMA), and utilize cosine distance as the metric for computing cost $C_{Appr}$ between the tracklet and detection appearance features. Note that the ReID components are not the focus of our paper.

\subsubsection{Algorithm Framework}
The association stage primarily consists of three stages: the first association stage for high-confidence objects, the second association stage for low-confidence objects (BYTE in ByteTrack), and the third association stage to recover lost tracklets with their last detection (OCR in OC-SORT).

Taking into account all the strong and weak cues, the final cost matrix basically comprises the following terms:
\begin{equation}
\label{eq:map5}
C = C_{HMIoU}+\lambda_{1}C_{Vel}+\lambda_{2}C_{Conf}+\lambda_{3}C_{Appr}
\end{equation}


\section{Experiments}
\subsection{Experimental Setting}
\subsubsection{Datasets}
We evaluated our design on various MOT benchmarks, including DanceTrack \cite{sun2022dancetrack}, MOT20 \cite{dendorfer2020mot20} and MOT17 \cite{milan2016mot16}. DanceTrack is currently one of the most challenging benchmarks in the MOT field, characterized by diverse non-linear motion patterns as well as frequent interactions and occlusions. It is noteworthy that the detection task in DanceTrack is relatively easy, making it an ideal benchmark to evaluate association performance. MOT20 was developed to evaluate algorithms under dense objects and severe occlusions. MOT17 is a widely used standard benchmark in MOT, in which the motion is mostly linear. Given the characteristics of these benchmarks, we primarily focus on comparing our method on DanceTrack as we aim to improve association performance with weak cues in challenging situations. We use MOT17 and MOT20 to evaluate the generalization ability of our method under diverse scenarios. The MOT17 validation set follows a widely adopted convention \cite{zhou2020tracking}, where the train set is split into halves for training and validation.

\subsubsection{Metrics}
We selected HOTA \cite{luiten2021hota} as our primary metric due to its higher-order nature. HOTA combines several sub-metrics that evaluate algorithms from different perspectives, providing a comprehensive assessment of algorithm performance. We also include other well-established metrics, such as MOTA \cite{bernardin2008evaluating} and IDF1 \cite{ristani2016performance}. IDF1 reflects the association aspect of the tracker, while MOTA is primarily influenced by detection performance.

\subsubsection{Implementation Details}
To ensure a fair comparison and demonstrate the superiority of our Hybrid-SORT, we directly adapt publicly available detection and ReID models from existing works. Specifically, for the detection part, we use the same detection model (i.e., YOLOX \cite{ge2021yolox}) as our baseline OC-SORT. Likewise, for the ReID part, we use the model (i.e., BoT \cite{luo2019strong}) in BoT-SORT \cite{aharon2022botsort}. The dimension of the appearance feature is 2048. The weight hyper-parameter of the confidence cost matrix in the first and second association stages are 1.5 and 1.0 on DanceTrack, 1.0 and 1.0 on other benchmarks. The weight of ROCM cost is 0.2, the same as OCM in OC-SORT. The IoU threshold to reject a match is set to 0.15 on DanceTrack, and 0.25 on other benchmarks. Following ByteTrack \cite{zhang2022bytetrack}, FPS is measured with FP16-precision \cite{micikevicius2018mixed} with batchsize of 1. The hardware is a single V100 GPU with Intel Xeon(R) Silver 4214R CPU @ 2.40GHz.                                                                                                                                                                                                                                                                                                                                                                                                                                                                                                                                                                                                                                                                                                                                                                                                                                                                                                                                                                                                                                                                                                                                                                                                                                                                                                                                                                                                                                                                                                                                                                                                                                                                                                                                                                                                                                                                                                                                                                                                                                                                                                                                                                                                                                                                                                                                                                                                                                                                                                                                                                                                                                                                                                                                                                                                                                                                                                                                                                                                                                                                                                                                                                                                                                                                                                                                                                                                                                                                                                                                                                                                                                                                                                                                                                                                                                                                                                                                                                                                                                                                                                                                                                                                                                                                                                                                                                                                                                                                                                                                                                                                                                                                                                                                                                                                                                                                                                                                                                                                                                                                                                                                                                                                                                                                                                                                                                                                                                                                                                                                                                                                                                                                                                                                                                                                                                                                                                                                                                                                                                                                                                                                                                                                                                                                                                                                                                                                                                                                                                                                                                                                                                                                                                                                                                                                                                                                                                                                                                                                                                                                                                                                                                                                                                                                                                                                                                                                                                                                                                                                                                                                                                                                                                                                                                                                                                                                                                                                                                                                                                                                                                                                                                                                                                                                                                                                                                                                                                                                                                                                                                                                                                                                                                                                                                                                                                                                                                                                                                                                                                                                                                                                                                                                                                                                                                                                                                                                                                                                                                                                                                                                                                                                                                                                                                                                                                                                                                                                                                                                                                                                                                                                                                                                                                                                                                                                                                                                                                                                                                                                                                                                                                                                                                                                                                                                                                                                                                                                                                                                                                                                                                                                                                                                                                                                                                                                                                                                                                                                                                                                                                                                                                                                                                                                                                                                                                                                                                                                                                                                                                                                                                                                                                                                                                                                                                                                                                                                                                                                                                                                                                                                                                                                                                                                                                                                                                                                                                                                                                                                                                                                                                                                                                                                                                                                                                                                                                                                                                                                                                                                                                                                                                                                                                                                                                                                                                                                                                                                                                                                                                                                                                                                                                                                                                                                                                                                                                                                                                                                                                                                                                                                                                                                                                                                                                                                                                                                                                                                                                                                                                                                                                                                                                                                                                                                                                                                                                                                                                                                                                                                                                                                                                                                                                                                                                                                                                                                                                                                                                                                                                                                                                                                                                                                                                                                                                                                                                                                                                                                                                                                                                                                                                                                                                                                                                                                                                                                                                                                                                                                                                                                                                                                                                                                                                                                                                                                                                                                                                                                                                                                                                                                                                                                                                                                                                                                                                                                                                                                                                                                                                                                                                                                                                                                                                                                                                                                                                                                                                                                                                                                                                                                                                                                                                                                                                                                                                                                                                                                                                                                                                                                                                                                                                                                                                                                                                                                                                                                                                                                                                                                                                                                                                                                                                                                                                                                                                                                                                                                                                                                                                                                                                                                                                                                                                                                                                                                                                                                                                                                                                                                                                                                                                                                                                                                                                                                                                                                                                                                                                                                                                                                                                                                                                                                                                                                                                                                                                                                                                                                                                                                                                                                                                                                                                                                                                                                                                                                                                                                                                                                                                                                                                                                                                                                                                                                                                                                                                                                                                                                                                                                                                                                                                                                                                                                                                                                                                                                                                                                                                                                                                                                                                                                                                                                                                                                                                                                                                                                                                                                                                                                                                                                                                                                                                                                                                                                                                                                                                                                                                                                                                                                                                                                                                                                                                                                                                                                                                                                                                                                                                                                                                                                                                                                                                                                                                                                                                                                                                                                                                                                                                                                                                                                                                                                                                                                                                                                                                                                                                                                                                                                                                                                                                                                                                                                                                                                                                                                                                                                                                                                                                                                                                                                                                                                                                                                                                                                                                                                                                                                                                                                                                                                                                                                                                                                                                                                                                                                                                                                                                                                                                                                                                                                                                                                                                                                                                                                                                                                                                                                                                                                                                                                                                                                                                                                                                                                                                                                                                                                                                                                                                                                                                                                                                                                                                                                                                                                                                                                                                                                                                                                                                                                                                                                                                                                                                                                                                                                                                                                                                                                                                                                                                                                                                                                                                                                                                                                                                                                                                                                                                                                                                                                                                                                                                                                                                                                                                                                                                                                                                                                                                                                                                                                                                                                                                                                                                                                                                                                                                                                                                                                                                                                                                                                                                                                                                                                                                                                                                                                                                                                                                                                                                                                                                                                                                                                                                                                                                                                                                                                                                                                                                                                                                                                                                                                                                                                                                                                                                                                                                                                                                                                                                                                                                                                                                                                                                                                                                                                                                                                                                                                                                                                                                                                                                                                                                                                                                                                                                                                                                                                                                                                                                                                                                                                                                                                                                                                                                                                                                                                                                                                                                                                                                                                                                                                                                                                                                                                                                                                                                                                                                                                                                                                                                                                                                                                                                                                                                                                                                                                                                                                                                                                                                                                                                                                                                                                                                                                                                                                                                                                                                                                                                                                                                                                                                                                                                                                                                                                                                                                                                                                                                                                                                                                                                                                                                                                                                                                                                                                                                                                                                                                                                                                                                                                                                                                                                                                                                                                                                                                                                                                                                                                                                                                                                                                                                                                                                                                                                                                                                                                                                                                                                                                                                                                                                                                                                                                                                                                                                                                                                                                                                                                                                                                                                                                                                                                                                                                                                                                                                                                                                                                                                                                                                                                                                                                                                                                                                                                                                                                                                                                                                                                                                                                                                                                                                                                                                                                                                                                                                                                                                                                                                                                                                                                                                                                                                                                                                                                                                                                                                                                                                                                                                                                                                                                                                                                                                                                                                                                                                                                                                                                                                                                                                                                                                                                                                                                                                                                                                                                                                                                                                                                                                                                                                                                                                                                                                                                                                                                                                                                                                                                                                                                                                                                                                                                                                                                                                                                                                                                                                                                                                                                                                                                                                                                                                                                                                                                                                                                                                                                                                                                                                                                                                                                                                                                                                                                                                                                                                                                                                                                                                                                                                                                                                                                                                                                                                                                                                                                                                                                                                                                                                                                                                                                                                                                                                                                                                                                                                                                                                                                                                                                                                                                                                                                                                                                                                                                                                                                                                                                                                                                                                                                                                                                                                                                                                                                                                                                                                                                                                                                                                                                                                                                                                                                                                                                                                                                                                                                                                                                                                                                                                                                                                                                                                                                                                                                                                                                                                                                                                                                                                                                                                                                                                                                                                                                                                                                                                                                                                                                                                                                                                                                                                                                                                                                                                                                                                                                                                                                                                                                                                                                                                                                                                                                                                                                                                                                                                                                                                                                                                                                                                                                                                                                                                                                                                                                                                                                                                                                                                                                                                                                                                                                                                                                                                                                                                                                                                                                                                                                           

\subsection{Benchmark Results}
In this section, we present benchmark results on DanceTrack, MOT20 and MOT17. Methods with identical detection results are grouped together at the bottom of each Table. 

We emphasize that Hybrid-SORT consistently outperforms the baseline OC-SORT in all three datasets with negligible additional computation and still maintains Simple, Online and Real-Time (SORT) characteristics, even though its performance lags slightly behind by a few works with much heavier models (i.e., MOTRv2), offline pipelines (i.e., SUSHI) or complex pipelines (i.e., MotionTrack and FineTrack) on certain datasets. 

The limited improvement of Hybrid-SORT on MOT17/20 largely attributes to the inherent shortcomings of the datasets themselves. Prominent studies such as DanceTrack \cite{sun2022dancetrack} and PersonPath22 \cite{shuai2022large} present two key arguments. First, the performance of methods may not be accurately assessed due to the limited sizes of MOT17/20, which are nearly 10$\times$ smaller than DanceTrack. Second, the two datasets mostly consist of simple linear motions and the performance becomes relatively saturated.

\subsubsection{DanceTrack}
Compared to the previous state-of-the-art heuristic tracker OC-SORT, Hybrid-SORT exhibits significantly superior performance (i.e., 7.6 HOTA), with identical association inputs and nearly identical computational complexity (refer to Table \ref{table:DanceTrack}). The results provide convincing evidence that the introduction and modeling of multiple types of weak cues, such as confidence state and height state, can effectively and efficiently resolve ambiguous and incorrect matches where strong cues fail. Further, with an independent ReID model, Hybrid-SORT-ReID achieves a state-of-the-art HOTA of 65.7 on DanceTrack for the heuristic tracker. For trackers with learnable matcher which show higher performance than Hybrid-SORT, MOTRv2 is also based on YOLOX detector but utilized a modified Deformable DETR \cite{zhu2020deformable} with 6 layers of Transformer encoder and 6 layers of Transformer decoder as the matcher, while SUSHI employs GNNs as the matcher with a totally offline pipeline.

\begin{table}[!t]
\centering
\small
{
\begin{tabular}{l|ccc} 

\hline\noalign{\smallskip}
Tracker & HOTA $\uparrow$ & IDF1 $\uparrow$ & MOTA $\uparrow$\\
\noalign{\smallskip}
\hline
\noalign{\smallskip}
\textit{\color{gray}{Learnable Matcher:}} & & & \\
MOTR  & 54.2 & 51.5 & 79.7\\
MOTRv2  & 69.9 & 71.7 & 91.9 \\
SUSHI  & 63.3 & 63.4 & 88.7\\
\hline
\textit{\color{gray}{Heuristic Matcher:}} & & &  \\
CenterTrack  & 41.8 & 35.7 & 86.8\\
FairMOT   & 39.7 & 40.8 & 82.2\\
QDTrack   & 45.7 & 44.8  & 83.0\\
FineTrack  & 52.7 & 59.8  & 89.9\\
\rowcolor{gray!10} SORT   & 47.9 & 50.8  & \textbf{91.8}\\
\rowcolor{gray!10} DeepSORT    & 45.6 & 47.9  & 87.8\\
\rowcolor{gray!10} ByteTrack    & 47.3 & 52.5  & 89.5 \\
\rowcolor{gray!10} GHOST    & 56.7 & 57.7  & 91.3 \\
\rowcolor{gray!10} OC-SORT   & 54.6 & 54.6  & 89.6\\
\rowcolor{gray!10} \textbf{Hybrid-SORT}   & 62.2 & 63.0  & 91.6\\
\rowcolor{gray!10} \textbf{Hybrid-SORT-ReID}   & \textbf{65.7} & \textbf{67.4}  & \textbf{91.8}\\
\hline
\end{tabular}
}
\caption{Results on DanceTrack test set. Methods in the gray block share the same detections. The highest-ranking  heuristic matcher is emphasized in bold.}
\label{table:DanceTrack}
\end{table}

\subsubsection{MOT20}
Hybrid-SORT achieves superior performance in the MOT20 test set (as shown in Table \ref{table:MOT20}) with high inference speed. Specifically, Hybrid-SORT surpasses OC-SORT in all metrics (i.e., 0.4 HOTA, 0.3 IDF1, and 0.9 MOTA), with practically indistinguishable additional computation. By utilizing an independent ReID model, Hybrid-SORT achieves a state-of-the-art performance of HOTA 63.9 on MOT20 for the heuristic tracker. The results demonstrate the effectiveness, robustness, and generalization of the proposed method in modeling weak cues for clustered and heavily occluded scenarios with dense objects.

\begin{table}
\centering
\small
{
\begin{tabular}{l|cccccccc}
\hline\noalign{\smallskip}
Tracker & HOTA $\uparrow$ & IDF1 $\uparrow$ & MOTA $\uparrow$ \\
\noalign{\smallskip}
\hline
\noalign{\smallskip}
\textit{\color{gray}{Learnable Matcher:}} \\
TrackFormer  & 54.7 & 65.7  & 68.6\\
MOTRv2  & 61.0& 73.1 & 76.2 \\
UTM   & 62.5 & 76.9  & 78.2\\
SUSHI  & 64.3 & 79.8 & 74.3\\
\hline
\textit{\color{gray}{Heuristic Matcher:}} \\
FairMOT  & 54.6  & 67.3& 61.8\\
CSTrack  & 54.0& 66.6 & 68.6 \\
FineTrack  & 63.6 & \textbf{79.0} & 77.9 \\
MotionTrack  & 62.8 & 76.5 & 78.0 \\
\rowcolor{gray!10} ByteTrack  & 61.3  & 75.2& 77.8\\
\rowcolor{gray!10} BoT-SORT   & 63.3 &  77.5 & 77.8 \\
\rowcolor{gray!10} GHOST   & 61.2  & 75.2 & 73.7 \\
\rowcolor{gray!10} OC-SORT   & 62.1 & 75.9& 75.5 \\
\rowcolor{gray!10} \textbf{Hybrid-SORT}  & 62.5 & 76.2 & 76.4\\
\rowcolor{gray!10} \textbf{Hybrid-SORT-ReID}  & 63.9 & 78.4 & 76.7 \\
\hline
\end{tabular}
}
\caption{Results on MOT20-test with the private detections. Methods in the gray block share the same detections. The highest-ranking  heuristic matcher is emphasized in bold.}
\label{table:MOT20}
\end{table}

\subsubsection{MOT17}
We present the performance of Hybrid-SORT on MOT17 in Table \ref{table:MOT17}. Specifically, Hybrid-SORT surpasses the previous state-of-the-art tracker OC-SORT in all metrics (i.e., 0.4 HOTA, 0.9 IDF1, and 1.3 MOTA) with negligible additional computation. By incorporating an independent ReID model, Hybrid-SORT further accomplishes performance improvements, setting a superior HOTA of 64.0 on MOT17. It is important to note that our method is primarily designed to address the challenges of object clustering and complex motion patterns. Nevertheless, even when applied to the MOT17 dataset, which represents a more general and easier scenario of linear motion patterns, our method consistently exhibits enhanced tracking performance.

\begin{table}[!ht]
\centering
\small
{
\begin{tabular}{l|ccc}
\hline\noalign{\smallskip}
Tracker & HOTA $\uparrow$ & IDF1 $\uparrow$ & MOTA $\uparrow$  \\
\noalign{\smallskip}
\hline
\noalign{\smallskip}
\textit{\color{gray}{Learnable Matcher:}}  & & & \\
TrackFormer  & 57.3& 68.0 & 74.1 \\
MOTR   & 57.8& 68.6 & 73.4 \\
MOTRv2  & 62.0& 75.0 & 78.6 \\
UTM    &  64.0 & 78.7& 81.8 \\
SUSHI  & 66.5 & 83.1& 81.1 \\
\hline
\textit{\color{gray}{Heuristic Matcher:}}  & & & \\
CenterTrack   & 52.2 & 64.7 & 67.8\\
QDTrack   & 53.9& 66.3 & 68.7 \\
FairMOT  & 59.3 & 72.3 & 73.7\\
CSTrack  & 59.3 & 72.6 & 74.9\\
FineTrack  & 64.3 & 79.5& 80.0 \\
MotionTrack   & \textbf{65.1} & 80.1 & \textbf{65.1} \\
\rowcolor{gray!10} ByteTrack    & 63.1 & 77.3 & 80.3 \\
\rowcolor{gray!10} BoT-SORT    & 65.0 & \textbf{80.2} & 80.5 \\
\rowcolor{gray!10} GHOST    & 62.8 & 77.1& 78.7 \\
\rowcolor{gray!10} OC-SORT    & 63.2  & 77.5& 78.0\\
\rowcolor{gray!10} \textbf{Hybrid-SORT}   & 63.6 & 78.4& 79.3 \\
\rowcolor{gray!10} \textbf{Hybrid-SORT-ReID}    & 64.0 & 78.7 & 79.9\\
\hline
\end{tabular}
}
\caption{Results on MOT17-test with the private detections. Methods in the gray block share the same detections. The highest-ranking  heuristic matcher is emphasized in bold.}
\label{table:MOT17}
\end{table}

\subsection{Ablation Study}
\subsubsection{Component Ablation}
As shown in Table \ref{table:component_ablation}. The results demonstrate the effectiveness and high efficiency of the proposed modules in Hybrid-SORT. The confidence state modeled by TCM significantly enhances the performance, with improvements of 4.0 HOTA. And notably, TCM only has a minor impact on inference speed (-0.7 FPS). Similarly, the utilization of height state by HMIoU leads to clear improvements in HOTA by 1.6 while barely affecting inference speed (-0.1 FPS). ROCM also enhances the association performance in HOTA by 0.6. However, ROCM reduces the inference speed by 1.5 FPS due to more temporal intervals and modeled points. With a commonly used ReID model, Hybrid-SORT-ReID further boosts the HOTA by 3.7, but the inference speed becomes near real-time. Note that the efficient incorporation of the ReID model into the MOT framework is beyond the scope of this paper.

\begin{table}
\centering
\small
{
\begin{tabular}{cccc|cc}
\noalign{\smallskip}
\hline\noalign{\smallskip}
ROCM & TCM & HMIoU & ReID & HOTA $\uparrow$ & FPS $\uparrow$ \\
\noalign{\smallskip}
\hline
\noalign{\smallskip}
   &   &  &  & 53.1  & \textbf{30.1} \\
\ding{51}&   &  &  & 53.7  & 28.6 \\
\ding{51}&\ding{51}&  &  & 57.7  & 27.9 \\
\ding{51}&\ding{51}&\ding{51}& & 59.3 & 27.8  \\
\ding{51}&\ding{51}&\ding{51}&\ding{51}& \textbf{63.0}  & 15.5  \\
\hline
\end{tabular}
}
\caption{Components ablation on DanceTrack-val. Consistent and significant improvements are observed using the proposed metrics TCM, HMIoU, and ROCM while maintaining real-time capacity.}
\label{table:component_ablation}
\end{table}

\begin{table}[t]
\centering
\small
{
\begin{tabular}{cc|ccc}
\hline\noalign{\smallskip}
first stage & second stage & HOTA $\uparrow$ & IDF1 $\uparrow$& MOTA $\uparrow$ \\
\noalign{\smallskip}
\hline
\noalign{\smallskip}
  -- &  -- & 53.7 & 53.2& 88.9  \\
Kalman  &  -- & 56.6 & 56.6& 89.2  \\
Kalman  &Kalman  & 57.3 & 57.9& 89.2 \\
Linear  &  -- & 52.6 & 52.1& 89.0  \\
Linear  &Linear  & 53.9 & 53.1& 89.2 \\
Kalman  &Linear  & \textbf{57.7}& \textbf{58.5} & \textbf{89.4} \\
\hline
\end{tabular}
}
\caption{Different confidence modeling on DanceTrack-val. The Kalman Filter is effective for unobstructed objects, while Linear Prediction is suitable for occluded objects.}
\label{table:confidence_modeling}
\end{table}

\begin{table}[t]
\centering
\small
{
\begin{tabular}{c|ccc}
\hline\noalign{\smallskip}
 & HOTA $\uparrow$  & IDF1 $\uparrow$ & MOTA $\uparrow$\\
\noalign{\smallskip}
\hline
\noalign{\smallskip}
IoU  & 57.7 & 58.5 & 89.4 \\
WMIoU   & 52.6 & 52.0 & 89.0 \\
HMIoU  & \textbf{59.3}  & \textbf{60.6} & \textbf{89.5} \\
\hline
\end{tabular}
}
\caption{Results of different IoU in DanceTrack-val. The regular height state provides benefits while the irregular width state causes harm.}
\label{table:different_iou}
\end{table}


\subsubsection{Modeling Strategies in TCM}
In Table \ref{table:confidence_modeling}, we investigate the performance of Kalman Filter and Linear Prediction for confidence state modeling on the DanceTrack-val. In the first association stage with high-confidence detections, Kalman Filter significantly boosts the association performance by 2.9 HOTA, while Linear Prediction decreases HOTA by 1.1. We attribute the results to the fact that high-confidence detections usually do not suffer from heavy occlusion, thus the confidence is stable and does not exhibit a clear directional trend. So Kalman Filter models the confidence state well but Linear Prediction fails. In the second association stage with low-confidence detections, both Kalman Filter and Linear Prediction perform well (0.7 and 1.1 HOTA, respectively). The confidence of occluded objects can decrease or increase rapidly depending on whether the clustering starts or ends. Kalman Filter is relatively incapable of modeling such sudden changes and the estimations usually lag behind the actual confidence. However, Linear Prediction can model the directional changes well.

\begin{table}[t]
\centering
\small
{
\begin{tabular}{c|c|cc}
\hline\noalign{\smallskip}
Tracker & TCM & DanceTrack & MOT17 \\
\noalign{\smallskip}
\hline
\noalign{\smallskip}
\multirow{2}{*}{ByteTrack}  &   & 47.06 & 67.85 \\
   &\ding{51}  & 49.32 (+2.3) & 68.03 (+0.2) \\
\hline
\multirow{2}{*}{SORT}  &   & 48.34 & 66.32 \\
 & \ding{51}  & 51.80 (+3.5) & 66.52 (+0.2)\\
\hline
\multirow{2}{*}{MOTDT}  &   & 36.47 & 65.32\\
  & \ding{51}  & 37.66 (+1.2) & 65.62 (+0.3)\\
\hline
\multirow{2}{*}{DeepSORT}  &   & 40.38 & 63.45\\
& \ding{51}  & 45.29 (+4.9) & 64.36 (+0.9)\\
\hline
\end{tabular}
}
\caption{TCM in other representative trackers. TCM consistently enhances tracking performance.}
\label{table:tcm_generalization}
\end{table}

\begin{table}[!ht]
\centering
\small
{
\begin{tabular}{c|c|cc}
\hline\noalign{\smallskip}
Tracker & HMIoU & DanceTrack & MOT17 \\
\noalign{\smallskip}
\hline
\noalign{\smallskip}
\multirow{2}{*}{ByteTrack} &   & 47.06 & 67.85 \\
 &\ding{51}  & 49.68 (+2.6) & 67.70 (-0.2) \\
\hline
\multirow{2}{*}{SORT}  &   & 48.34 & 66.32 \\
 & \ding{51}  & 49.96 (+1.6) & 67.30 (+1.0)\\
\hline
\multirow{2}{*}{MOTDT}  &   & 36.47 & 65.32\\
 & \ding{51}  & 36.83 (+0.4) & 65.21 (-0.1)\\
\hline
\multirow{2}{*}{DeepSORT} &   & 40.38 & 63.45\\
 & \ding{51}  & 41.23 (+0.9) & 63.64 (+0.2)\\
\hline
\end{tabular}
}
\caption{HMIoU in other representative trackers. HMIoU consistently enhances tracking performance.}
\label{table:vhiou_generalization}
\end{table}

\subsubsection{Height State or Width State}
We argue the height state, rather than the width state, can benefit association. Similar to the HMIoU, We propose Width Modulated IoU (WMIoU) by replacing height with width. As shown in Table \ref{table:different_iou}, width state significantly hurt association performance, whereas the height state is beneficial. The reason is the box width varies irregularly due to pose changes or limb movements, posing a challenge for precise estimation by the Kalman Filter. In contrast, the height state undergoes relatively short and continuous changes during squatting or standing up, making it effectively modeled by the Kalman Filter.


\subsubsection{Generality on Other Trackers}
We applied our design to other 4 representative heuristic trackers, namely SORT \cite{bewley2016simple}, DeepSORT \cite{wojke2017simple}, MOTDT \cite{chen2018real}, and ByteTrack \cite{zhang2022bytetrack}. Among these trackers, SORT, and ByteTrack rely solely on spatial information, while MOTDT and DeepSORT jointly utilize both spatial and appearance information. The results are presented in Table \ref{table:tcm_generalization} and Table \ref{table:vhiou_generalization}, where a significant improvement can be observed in both DanceTrack and MOT17 datasets for all aforementioned trackers. For instance, our design TCM improves DeepSORT by 4.9 HOTA in DanceTrack and 0.9 HOTA in MOT17, while our HMIoU boosts SORT by 1.6 HOTA in DanceTrack and 1.0 HOTA in MOT17. These results provide convincing evidence that our insight of introducing weak cues like confidence state and height state as compensation for strong cues is effective and generalizes well across different trackers and scenarios. Moreover, our method can be readily applied to existing trackers in a plug-and-play and training-free manner for enhanced performance.

\section{Conclusion}
In this paper, we demonstrate that the common and long-standing challenge of heavy occlusion and clustering can be effectively and efficiently alleviated with previously overlooked weak cues (e.g. confidence state, height state, and velocity direction). These weak cues can compensate for the limitations of strong cues. Then, we propose Hybrid-SORT by introducing simple modeling for the newly incorporated weak cues and leveraging both strong and weak cues, which significantly improves the association performance. Furthermore, Hybrid-SORT still maintains Simple, Online and Real-Time (SORT) characteristics, and can be readily applied to existing trackers in a plug-and-play and training-free way. Extensive experiments demonstrate the strong generalization ability of Hybrid-SORT across diverse trackers and scenarios. With widely used appearance information, Hybrid-SORT achieves superior performance over state-of-the-art methods, with a much simpler pipeline and faster association. We hope that the aforementioned characteristics of Hybrid-SORT make it attractive for diverse scenarios and devices with limited computational resources.

\section*{Acknowledgments}
The paper is supported in part by National Natural Science Foundation of China (Nos.U23A20384, 62293542), in part by Talent Fund of Liaoning Province (XLYC2203014), and in part by Fundamental Research Funds for the Central Universities (No.DUT22QN228).

\bibliography{aaai24}

\end{document}